\documentclass[10pt,twocolumn,letterpaper]{article}

\usepackage{iccv}
\usepackage{times}
\usepackage{epsfig}
\usepackage{graphicx}
\usepackage{amsmath}
\usepackage{amssymb}
\usepackage{mathtools}
\usepackage{bbm}
\usepackage{caption}
\usepackage{subcaption}

\graphicspath{{images},{prebuiltimages}}
\usepackage[titlenumbered,ruled,noend,vlined,norelsize,slide]{algorithm2e}

\SetCommentSty{mycommfont}
\SetKwInput{KwInput}{Input}
\usepackage{shortcuts_iccv}
\usepackage[scientific-notation=false,group-minimum-digits=4]{siunitx}

\usepackage[pagebackref=true,breaklinks=true,letterpaper=true,colorlinks,bookmarks=false]{hyperref}
\hypersetup{
    colorlinks = true,
    linkcolor = blue,
    anchorcolor = blue,
    citecolor = cyan,
    filecolor = blue,
    urlcolor = blue
    }
\usepackage{adjustbox}

\usepackage[capitalize,noabbrev]{cleveref}
\usepackage{amsfonts}

 \iccvfinalcopy 

\def\iccvPaperID{12504} 

\ificcvfinal\fi

\begin{document}

\title{A two-head loss function for deep Average-K classification}

\author{Camille Garcin$^{1, 2}$, Maximilien Servajean$^{3, 4}$, Alexis Joly$^{2, 3}$, Joseph Salmon$^{1, 5}$\\
$^{1}$IMAG, Univ Montpellier, CNRS, Montpellier, France  \\
$^{2}$Inria, LIRMM, Univ Montpellier, CNRS, Montpellier, France \\
$^{3}$LIRMM, Univ Montpellier, CNRS, Montpellier, France \\
$^{4}$AMIS, Paul Valery University, Montpellier, France \\
$^{5}$Institut Universitaire de France (IUF) \\
{\tt\small camille.garcin@inria.fr, maximilien.servajean@lirmm.fr} \\
{\tt\small alexis.joly@inria.fr, joseph.salmon@umontpellier.fr}
}

\maketitle
\ificcvfinal\thispagestyle{empty}\fi


\begin{abstract}
    Average-$K$ classification is an alternative to top-$K$ classification in which the number of labels returned varies with the ambiguity of the input image but must average to $K$ over all the samples.
    A simple method to solve this task is to threshold the softmax output of a model trained with the cross-entropy loss.
    This approach is theoretically proven to be asymptotically consistent, but it is not guaranteed to be optimal for a finite set of samples.
    In this paper, we propose a new loss function based on a multi-label classification head in addition to the classical softmax.
    This second head is trained using pseudo-labels generated by thresholding the softmax head while guaranteeing that $K$ classes are returned on average.
    We show that this approach allows the model to better capture ambiguities between classes and, as a result, to return more consistent sets of possible classes.
    Experiments on two datasets from the literature demonstrate that our approach outperforms the softmax baseline, as well as several other loss functions more generally designed for weakly supervised multi-label classification.
    The gains are larger the higher the uncertainty, especially for classes with few samples.
\end{abstract}



\section{Introduction}
\label{sec:Introduction}
\begin{figure}[t]
  \centering
  \includegraphics[width=0.99\linewidth]{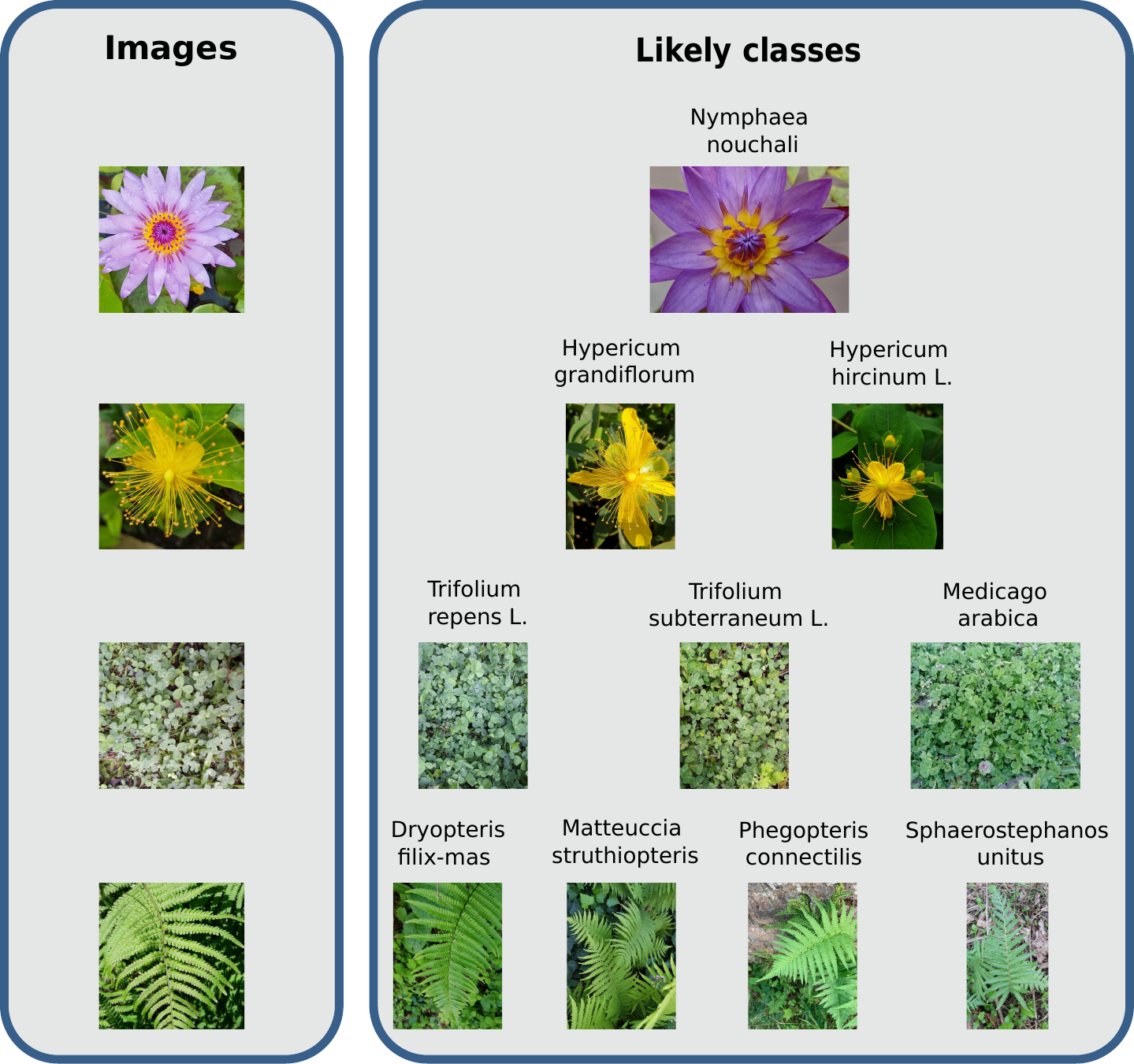}
  \caption{Variety of ambiguity in Pl@ntNet-300K dataset \cite{plantnet-300k}. The images on the left are from the dataset while the ones on the right represent the likely corresponding classes.
    The first row corresponds to the case where there is almost no uncertainty, the second row to the case where two classes are viable candidates for the image, etc.}
  \label{fig:plant_pictures}
\end{figure}
The rise of visual sensors and the democratization of crowd-sourcing approaches (\eg citizen science) are contributing to the emergence of new massive image datasets with a very large number of classes \cite{inat2017, plantnet-300k, yuan2021eproduct, robertson2019training}.
These datasets contain many classes that are similar to each other and are prone to label ambiguity due to the crowd-sourced acquisition and/or annotation process.
In such cases, it is difficult to achieve high levels of top-1 accuracy.
This is consistent with theory: most of the time, the Bayes classifier has a non-zero error rate due to the random nature of the underlying generation process \cite{der2009aleatory}.
This is problematic for systems that aim to provide their users with a correct answer from an image \cite{hart2023assessing, machado2021systematic}.

One way to deal with this difficulty is to allow the classifier to return more than one class.
The most standard way to do this is through top-$K$ classification, in which we allow the classifier to return exactly $K$ candidate classes for all images \cite{lapin2015}.
For instance, top-$K$ accuracy is the official ImageNet \cite{imagenet} metric.

Although top-$K$ classifiers reduce the error rate by returning $K$ classes, they lack flexibility: for some clean, unambiguous images, it is not necessary to return $K$ candidate classes.
Conversely, for some ambiguous images, $K$ may not be sufficient for the true class to be in the returned set.
\Cref{fig:plant_pictures} illustrates how ambiguity can vary from one image to the other.

To address this variability, classifiers that return a variable number of classes from an input must be used.
There exists a broad range of set-valued classifiers~\cite{chzhen2021set} and strategies ~\cite{fontana2023conformal}, some more flexible than others.
In this paper, we focus on average-$K$ classifiers \cite{denis} that return class sets of average size $K$, where the average is taken over the dataset.
This constraint is less strict than for top-$K$ classification, where the cardinal of the returned sets must be $K$ for each instance.
This flexibility allows returning more than $K$ classes for ambiguous images and less than $K$ classes for easy ones.

Controlling the average number of classes returned is useful for many applications, as it meets both UI (User Interface) design needs (\eg the average number of results should fit on a mobile app screen) and UX (User eXperience) design needs (\eg a recommender system should not recommend too many items on average).
In some cases, the expected average size may also be known or estimated from other data sources (\eg the average number of species present at a given location).

The simplest approach to perform average-$K$ classification is to threshold the softmax predictions of a deep neural network optimized with cross-entropy loss.
While cross-entropy is theoretically grounded in the infinite sample limit \cite{lorieulthese}, no guarantee exists in the finite sample case.

In this paper, we propose a novel method to optimize average-$K$ accuracy by adding an auxiliary head to a classical deep learning model.
One head is responsible for identifying candidate classes to be returned in the set and the second head maximizes the likelihood of each candidate class.
We experiment on two datasets and report significant improvements over vanilla cross-entropy training and other methods.

\section{Problem statement}
\label{sec:problem_statement}
We are in the multi-class setting, where the image/label pairs $(X, Y)$ are assumed to be generated \iid by an unknown probability distribution $\bbP$.
Each image $x$ is associated with a label $y \in [L] \coloneqq \{1, \dots, L \}$, where $L$ is the number of classes.
Traditional multi-class classifiers are functions $f: \cX \rightarrow [L]$ that map an image to a single label.

To reduce the risk of not returning the true class associated to an image, we are interested in set-valued classifiers $g: \cX \rightarrow 2^{[L]}$ that map an image to a set of labels \cite{chzhen2021set}, where $2^{[L]}$ is the power set of $[L]$.
Our objective is then to build a classifier $g$ with minimal risk $\mathcal{R}(g) \coloneqq \bbP(Y \notin g(X))$.

A trivial solution would be to take $g(x) = [L]$ for all $x \in \cX$, \ie the classifier that returns the set of all classes for each input image.
To build a useful classifier, a constraint must be enforced on this optimization problem (as discussed in \cite{chzhen2021set} where a unified framework encompassing several possible constraints is proposed).
In this paper, we focus on the average set size constraint that controls the average size of the returned set:
\begin{align}
    \bbE_{X}[|g(X)|] \leq K \enspace,   (\mathcal{C})
\end{align}
where $K$ is an integer and $|\cdot|$ denotes the cardinal.

It has been shown \cite{denis} that the optimal classifier $g^{*}$ minimizing $\mathcal{R}$ while satisfying $\mathcal{C}$ is:
\begin{align}
    g^{*}(x) = \{j \in [L], \bbP(Y = j | X=x) \geq \lambda \} \enspace,
\end{align}
where $\lambda \in [0, 1]$ is calibrated so that $K$ classes are returned on average.

These quantities are theoretical since we do not know $\bbP$.
In practice, given a real dataset, the typical workflow for building an average-$K$ classifier is to learn the model with cross-entropy on the training set and compute the threshold --- so that on average $K$ classes are returned --- on the estimated conditional probabilities (softmax layer) with a validation set.
The model is then evaluated on the test set.

While cross-entropy is theoretically grounded in the infinite limit sample case \cite{lorieulthese}, there is no guarantee in the finite sample regime.
In this paper, we propose an alternative to cross-entropy to optimize average-$K$ accuracy by formulating the problem as a multi-label problem.


\section{Related work}
\label{sec:related_work}
Several works have studied set-valued classification.
The most studied case is top-$K$ classification.
In \cite{lapin2015}, the authors propose a top-$K$ hinge loss and optimize it with Stochastic Dual Coordinate Ascent \cite{sdca}.
In \cite{topk_yang}, the authors propose a slight variation of the top-$K$ hinge loss with stronger theoretical guarantees.
In \cite{berrada} and \cite{topk_icml2022}, the authors propose to smooth top-$K$ hinge losses to make them suitable for deep learning.

Average-$K$ classification is less studied.
In \cite{denis}, the authors derive the average-$K$ Bayes classifier.
They also show that minimizing a certain $\phi$-risk with specific conditions on $\phi$ implies minimizing the average-$K$ error.
In \cite{lorieulthese}, the authors show that strongly proper losses are consistent for average-$K$ classification,
which means that minimizing the expected value of such a loss leads to minimizing the expected value of the average-$K$ error.
They also show that the cross-entropy loss is strongly proper, which gives a theoretical argument for using the cross-entropy loss for average-$K$ classification.
While the two previous works give theoretical results in the infinite limit case, we instead propose a practical method for optimizing average-$K$ accuracy and show that it performs better than cross-entropy and other methods on real-world datasets.

Another framework close to set-valued classification is conformal prediction \cite{vovk2005algorithmic, shafer2008tutorial, fontana2023conformal}.
Conformal prediction also aims to generate prediction sets, but it relies on the use of calibration data to determine them based on hypothesis testing.
It has the advantage of working on any pre-trained classifier $\hat{f}: \cX \rightarrow [L]$, but it does not optimize the model itself towards the set-valued classification objective.
Furthermore, the availability of calibration data can be problematic, especially for classes with few training examples (often the most numerous in the case of real-world datasets \cite{plantnet-300k}).

Set-valued classification can also be connected to multi-label classification \cite{dembczynski2012label, liu2021emerging}.
Indeed, in both settings, a set of labels is predicted in output.
A crucial difference, however, lies in the data generation process.
In multi-label classification, each input $x$ is associated with a set of labels, whereas in the set-valued setting, each input $x$ is associated with a single label.
The work of \cite{cole2021} is the closest to ours.
The authors study the positive-only multi-label classification case, which is a particular case of multi-label classification where a single positive class is observed for the training images, but all the other classes are not observed (meaning they could be either positive or negative, we do not know).
Their objective is then to learn a good classifier with this partial information.
Our setting is different in the sense that in multi-class classification a single object is present in the image, but we want to return a set of possible classes to reduce the risk.
However, both settings share similarities: a single class label is available during training and for each input image we return a set of classes.

\section{Set-valued classification as multi-label}
\label{sec:multi_label}

\subsection{Preliminaries}
\label{subsec:preliminaries}

\begin{figure*}[t]
    \centering
    \includegraphics[width=0.99\linewidth]{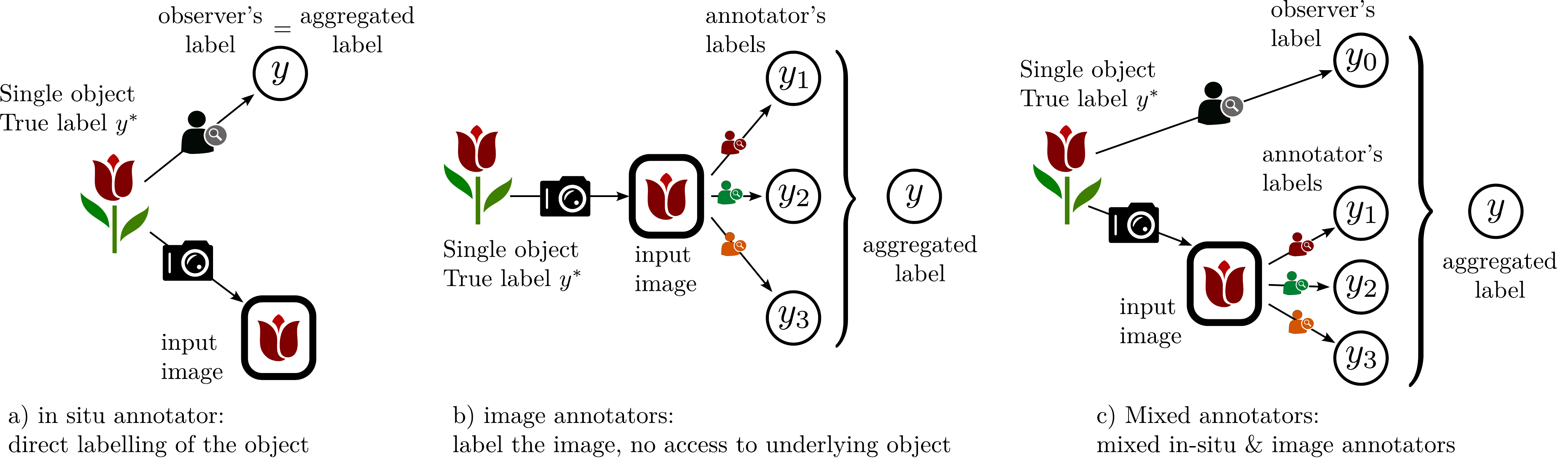}
    \caption{Common annotation processes for multi-class datasets. The label $y$ given to the input image is either obtained: a) after direct observation of the whole original object \eg \cite{little2020algorithm}
        b) by aggregating the labels of annotators having access only to a partial representation of the object (the input image), \eg \cite{cifar100, imagenet}
        c) with a combination of both previous cases \eg \cite{plantnet-300k}. In all cases, annotations errors can occur resulting in a final label $y$ different than the true class $y^{*}$.}
    \label{fig:datasets_labels}
\end{figure*}

In most multi-class datasets \cite{plantnet-300k, cifar100, imagenet}, the label $y$ associated with an object $o$ of true class $y^{*}$ is estimated by the annotators from a partial representation of $o$: an image $x$.
\Cref{fig:datasets_labels} summarizes how the final label $y$ is obtained in most cases.

It may be difficult to determine the true class $y^{*}$ of an object $o$ given the partial representation $x$.
The image may be of poor quality, lack a discriminative feature of $y^{*}$, etc.
In such cases, several candidate classes are plausible for the object $o$, which we denote by the set $\mathcal{S}(x) \in 2^{[L]}$, see \Cref{fig:plant_pictures}.

Regardless of this difficulty, a single label $y$ ---possibly different from $y^{*}$--- is selected in multi-class datasets.
In the case of multiple annotators, the default policy is to select the label with the most votes \cite{khattak_toward_2017}.

In this paper, we propose to dynamically estimate the sets $\mathcal{S}(x)$ during training and use $\mathcal{S}(x)$ as the ground truth labels with the binary cross-entropy loss in a multi-label fashion.

\subsection{Notation}
\label{subsec:notations}

Let $\mathcal{N}_{train}$, $\mathcal{N}_{val}$ and $\mathcal{N}_{test}$ denote respectively the training set, the validation set and the test set indices.
In the rest of the paper, we use mini-batch gradient descent \cite{ruder2016overview} to optimize all models.
In the following, $B \subset \mathcal{N}_{train}$ denotes a random batch of training input.

For $i \in \mathcal{N}_{train} \cup \mathcal{N}_{val} \cup \mathcal{N}_{test}$, $\mathbf{z}_{i} \in \bbR^{L}$ denotes the score vector ---logit--- predicted by the model, $z_{ij}$ its $j$-th component, and $y_{i} \in [L]$ the label assigned to $i$.

Given a batch $B = \{i_{1}, i_{2}, \dots, i_{|B|}\}\subset \mathcal{N}_{train}$, we define:
\begin{align*}
    Z^{B} =
    \begin{bmatrix}
        \text{---} \, \mathbf{z}_{i_1}    \, \text{---} \\
        \text{---} \, \mathbf{z}_{i_2}    \, \text{---} \\
        \dots                                           \\
        \text{---} \, \mathbf{z}_{i_{|B|}} \,  \text{---}
    \end{bmatrix} \quad \text{ and } \quad
    Y^{B} =
    \begin{bmatrix}
        y_{i_1} \\
        y_{i_2} \\
        \dots   \\
        y_{i_{|B|}}
    \end{bmatrix} \enspace,
\end{align*}
respectively the batch predictions and batch labels.

Let $P$ be a vector and $k \in \bbN^{*}$ a positive integer. $P_{[k]}$ will denote the $k$-th largest value of $P$.

Finally, let us note $\varsigma: \bbR^{L} \rightarrow \bbR^{L}$ the softmax function whose $j$-th component is given, for any $\mathbf{z}_{i} \in \bbR^{L}$ by:
\begin{align*}
    \varsigma_{j}(\mathbf{z}_{i}) = \frac{e^{z_{ij}}}{\sum_{k=1}^{L} e^{z_{ik}}} \enspace,
\end{align*}
and $\sigma$ the sigmoid function, defined for any $t\in\bbR$ by:
\begin{align*}
    \sigma (t)  = \frac {1}{1+e^{-t}} \enspace.
\end{align*}
\subsection{Cross-entropy loss}
\label{subsec:cross_entropy}
Given a logit vector $\mathbf{z}_{i} \in \bbR^{L}$ and a label $y_i$, the cross-entropy loss for example $i$ writes:
\begin{align}
    \ell_{\rm CE}(\mathbf{z}_{i}, y_{i})
     & = - \log(\varsigma_{y_{i}}(\mathbf{z}_{i})) \enspace.
\end{align}
The partial derivatives read:
\begin{align}
    \frac{\partial \ell_{\rm CE}}{\partial z_{j}}(\mathbf{z}_{i}, y_{i}) = \begin{cases}
                                                                               \varsigma_{y_{i}}(\mathbf{z}_{i}) (1 - \varsigma_{y_{i}}(\mathbf{z}_{i})) , & \text{if}\ j=y_{i} \\
                                                                               - \varsigma_{y_{i}}(\mathbf{z}_{i}) \cdot \varsigma_{j}(\mathbf{z}_{i}),    & \text{o.w.}
                                                                           \end{cases}\enspace,
\end{align}
which is positive only if $j = y_{i}$. \\

Therefore, after a gradient descent update on $\ell_{\rm CE}$, $z_{i,y_{i}}$ will increase and all other scores $(z_{ij})_{j \neq y_{i}}$ will decrease.

As stated in \Cref{sec:Introduction,sec:problem_statement,sec:related_work}, in the infinite limit case, $\ell_{\rm CE}$ has theoretical grounds \cite{lorieulthese}.
However, it is not clear that this approach is optimal when only scarce/noisy data is available.
Indeed, let us consider the case where two labels $y_i$ and $\tilde{y}_i$ are equiprobable for the image $x_{i}$:
\begin{align*}
    \bbP(Y = y_i | X=x_{i})=\bbP(Y = \tilde{y}_i | X=x_{i})=0.5 \enspace,
\end{align*}
and assume that the label $y_i$ was assigned to $x_{i}$ when the dataset was constructed.
With $\ell_{\rm CE}$, the score $z_{i, \tilde{y}_i}$ will decrease and $z_{i, {y}_i}$ will increase during training, while we would like both $z_{i, \tilde{y}_i}$ and $z_{i, {y}_i}$ to increase.
Hence, in the context of high ambiguity, it is reasonable to formulate the problem as a multi-label classification task, in which each image $x_{i}$ is associated with a set of labels $\mathcal{S}(x_{i}) \in 2^{[L]}$, the difficulty being that only one of them is observed, $y_{i}$.
\subsection{Assume negative}
\label{subsec:assume_negative}
A first multi-label approach similar to cross-entropy is to consider that for any $i \in \mathcal{N}_{train}$, $\mathcal{S}(x_{i}) = \{y_{i} \}$, \ie there is no ambiguity in the labels.
This results in the Assume Negative loss $\ell_{\rm AN}$ \cite{cole2021}, essentially the binary cross entropy-loss with a single positive label:
\begin{align}
    \label{eq:lan}
    \ell_{\rm AN}(\mathbf{z}_{i}, y_{i}) = - \sum_{j=1}^{L} \Big[ & \mathbbm{1}_{[j = y_{i}]} \log(\sigma(z_{ij}))                        \nonumber \\
    + \tfrac{1}{L-1}                                              & \mathbbm{1}_{[j \neq y_{i}]}  \log(1 - \sigma(z_{ij}))\Big] \enspace,
\end{align}
where the negative labels are weighted to have the same contribution as the positive label.
\subsection{Expected positive regularization}
\label{subsec:epr}
The problem with $\ell_{\rm AN}$ is that the second term of \Cref{eq:lan} assumes that the scores of all classes different from $y_i$ must be minimized, regardless of their relevance to example $i$.
Removing the second term yields the positive cross-entropy loss \cite{cole2021}:
\begin{align}
    \ell^{+}_{\rm BCE}(\mathbf{z}_{i}, y_{i}) = - \sum_{j=1}^{L} \mathbbm{1}_{[j=y_{i}]} \log(\sigma(z_{ij})) \enspace.
\end{align}
However, $\ell^{+}_{\rm BCE}$ is minimized by predicting 1 for all classes,
which is not desirable as it would lead to a large number of false positives.

A workaround proposed in \cite{cole2021} is to constrain the current batch probability predictions $\sigma(Z^{B})$ to sum to $K$ on average, where $\sigma$ is applied pointwise and the average is over the batch dimension.
Here $K$ is a hyperparameter that can be thought of as the expected average set size.
More formally, the expected number of positive classes can be estimated as:
\begin{align}
    \hat{K}(Z^{B}) = \frac{1}{|B|} \sum_{i \in B} \sum_{j=1}^{L} \sigma(z_{ij}) \enspace.
\end{align}
The Expected positive regularization loss \cite{cole2021} $\ell_{\rm EPR}$ then reads:
\begin{align}
    \label{eq:epr}
    \ell_{\rm EPR}(Z^{B}\!\!, Y^{B}) = \tfrac{-1}{|B|} \sum_{\mathclap{i \in B}} \log(\sigma(z_{i, y_{i}}))+\beta (\hat{K}(Z^{B}){-}K)^{2} , \raisetag{5pt}
\end{align}
where $\beta$ is a hyperparameter to be tuned on a validation set.

Although the idea behind $\ell_{\rm EPR}$ seems reasonable, there is an infinite number of combinations for the matrix $\sigma(Z^{B})$ to sum to $K$ on average.
In particular, the model could learn to place diffuse probabilities on all classes without promoting strong class candidate alternatives to the true label class in the training set.
\subsection{Online estimation of labels}
\label{subsec:role}
In \cite{cole2021}, the authors introduce a loss $\ell_{\rm ROLE}$ that builds on $\ell_{\rm EPR}$.
In addition, they keep a matrix estimate of the unobserved labels $\Theta \in \bbR^{n_{train} \times L}$, where $n_{train} {\coloneqq} |\mathcal{N}_{train}|$ is the number of training examples in the dataset.
During training, the labels predicted by the model are trained to match the current estimates of the labels in $\Theta$ via a cross-entropy term and, in addition, to satisfy the constraint in \Cref{eq:epr}.
The role of the predicted and estimated labels is then reversed.
For more details, we refer the reader to \cite{cole2021}.

Although $\ell_{\rm ROLE}$ is more sophisticated than EPR, we find in our experiments that it does not perform well.
Moreover, it requires tuning several hyperparameters and, most importantly, keeping in GPU memory a matrix of size $n_{train} \times L$, which is prohibitive for large datasets.

\section{Proposed method}
\label{sec:prop_method}
\begin{figure*}[t]
    \centering
    \includegraphics[width=0.99\linewidth]{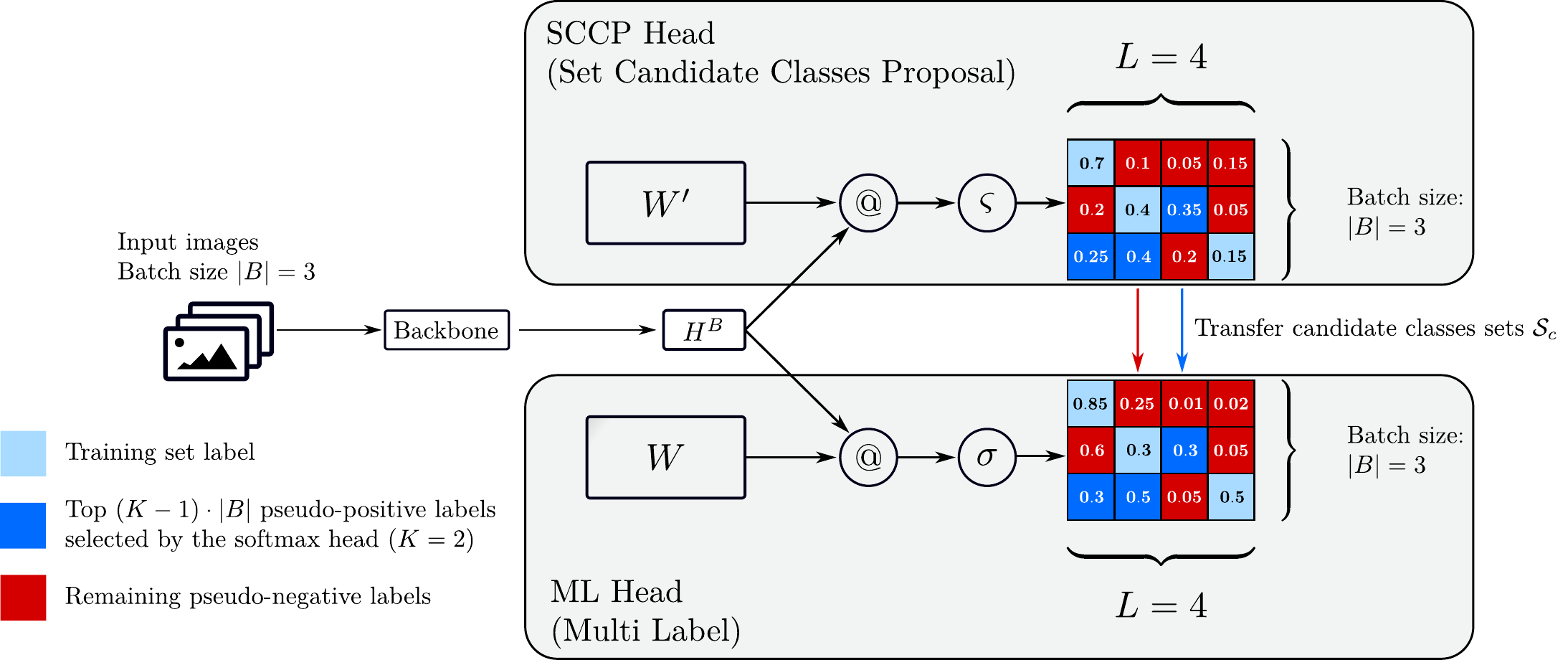}
    \caption{The Set Candidate Classes Proposal -SCCP- head (on top) is responsible for determining, for each example of the batch, which classes to include in the target label set.
        The light blue cells correspond to the training set labels. Only one label is assigned per example/row. The dark blue cells correspond to the classes selected as pseudo-positives by the SCCP head.
        They correspond to the top $(K-1)|B|$ highest values of the SCCP's softmax prediction matrix deprived of the light blue true labels.
        In this example, $K=2$ and $|B|=3$, so $(2-1)\times3 = 3$ classes are selected as pseudo-positives in the batch.
        They are assigned a pseudo-label 1 in the ML head. The remaining cells, the red ones, are those that were not selected as pseudo-positives by the SCCP head and are considered pseudo-negatives by the ML head (with a pseudo-label 0).
        Here @ denotes matrix multiplication.
    }
    \label{fig:schema_iccv}
\end{figure*}
\subsection{Outline}
\label{subsec:outline}
Let us assume that given a random batch $B \subset \mathcal{N}_{train}$ of examples, we are able to estimate for each $i \in B$ a set of possible classes different than $y_{i}$ which we denote $\mathcal{S}_{c}(x_{i})$.
We can then define $\mathcal{S}(x_{i}) = \{y_{i}\} \cup \mathcal{S}_{c}(x_{i})$.
A legitimate objective is to increase the scores $(z_{ij})_{j \in \mathcal{S}(x_{i})}$ and decrease the scores $(z_{ij})_{j \notin \mathcal{S}(x_{i})}$, which is achieved by the following binary cross-entropy loss:
\begingroup
\addtolength{\jot}{0.4em}
\begin{align}
    \label{eq:ml_head}
    \ell(Z^{B}\!\!, Y^{B}\!\!, \mathcal{S}^{B}_{c}) = & - \frac{1}{|B|}                                      \sum\limits_{\substack{i \in B \\ j \in [L]}}  \mathbbm{1}_{[j = y_{i}]} \log(\sigma(z_{ij}))           \\
                                                      & - \alpha\frac{\sum\limits_{\substack{i \in B                                        \\ j \in [L]}} \mathbbm{1}_{[j \in \mathcal{S}_{c}(x_{i})]} \log(\sigma(z_{ij}))}{\sum\limits_{\substack{i \in B \\ j \in [L]}} \mathbbm{1}_{[j \in \mathcal{S}_{c}(x_{i})]}}     \nonumber \\
                                                      & - \alpha\frac{\sum\limits_{\substack{i \in B                                        \\ j \in [L]}} \mathbbm{1}_{[j \notin \mathcal{S}(x_{i})]} \log(1 - \sigma(z_{ij}))}{\sum\limits_{\substack{i \in B \\ j \in [L]}} \mathbbm{1}_{[j \notin \mathcal{S}(x_{i})]}} \enspace, \nonumber
\end{align}
\endgroup

where the hyperparameter $\alpha \in \bbR^{+}$ controls the weighting between the training set observed labels and the candidate labels and can be seen as setting a different learning rate for the ``hypothetical" labels (\ie the pseudo-labels) and the observed labels.
We used the notation $\mathcal{S}^{B}_{c} = \{\mathcal{S}_{c}(x_{i}), i \in B\}$.
The main difficulty is: how to obtain the candidate labels $\mathcal{S}_{c}(x_{i})$?
\subsection{Candidate classes estimation}
\label{subsec:candidate_classes}
To this end, we propose a two-head model, where the first head is responsible for identifying the candidate classes $\mathcal{S}_{c}(x_{i})$ for each $i \in B$ (Set Candidate Classes Proposal -SCCP- head) and the second head (Multi-Label -ML- head) optimizes its predictions with the loss from \Cref{eq:ml_head} in a multi-label fashion with the candidate classes $\mathcal{S}_{c}(x_{i})$ estimated by the SCCP head.

Let us denote $H^{B} \in \bbR^{B \times d}$ as the output of the second to last layer of a deep neural network, where $d$ is the dimension of the latent space.
We define each head prediction as the output of a single linear layer building on $H^{B}$: $Z^{B} = H^{B}W \in \bbR^{B \times L}$ for the BCE head and $Z'^{B} = H^{B}W' \in \bbR^{B \times L}$ for the SCCP head.
To identify relevant candidate classes, we rely on cross-entropy for its theoretical foundations \cite{lorieulthese}.
More formally, we optimize the SCCP head with:
\begin{align}
    \ell_{\rm CE}(Z'^{B}, Y^{B}) = - \frac{1}{|B|} \sum_{i \in B} \log(\varsigma_{y_i}({\mathbf{z}}_{i}')) \enspace,
\end{align}

where ${\mathbf{z}}_{i}' \in \bbR^{L}$ is the score prediction of the SCCP head for example $i$.

To propose sets of candidate classes, we select the maximum activations of the matrix $[\Sigma^{-\infty}_{ij}]_{i \in B, j \in [L]}$ defined as:

\begin{align}
    \Sigma^{-\infty}_{ij} =  \begin{cases}
                                 \varsigma_{j}({\mathbf{z}}_{i}'), & \text{if}\ y_{i} \neq j \\
                                 -\infty,                          & \text{otherwise}
                             \end{cases} \enspace.
\end{align}

To construct the candidate classes sets $\mathcal{S}^{B}_{c}$, we then select the top $(K-1)|B|$ values of $\Sigma^{-\infty}$ to obtain sets of average size $K$.
More formally, for $i \in B$, we define $\mathcal{S}_{c}(x_{i})$ as:
\begin{align}
    \mathcal{S}_{c}(x_{i}) = \{j, \Sigma^{-\infty}_{ij} \: \text{is in the} \: \mathrm{top}\text{-}{\scriptstyle   (K-1)|B|} \: \text{values of} \: \Sigma^{-\infty}\} \enspace. \raisetag{-3pt}
\end{align}
This choice leads to sets of average size $K$ on the batch:
\begin{align}
    \frac{1}{|B|} \sum_{i \in B} |\mathcal{S}(x_{i})| & = \frac{1}{|B|} \sum_{i \in B} |\mathcal{S}_{c}(x_{i}) \cup \{y_i\}| \\
                                                      & = \frac{(K-1)|B| + |B|}{|B|} \nonumber                               \\
                                                      & = K                    \nonumber
\end{align}
An illustrative schema of the method is available in \Cref{fig:schema_iccv}.
We can now plug the estimated candidate sets $\mathcal{S}^{B}_{c}$ into \Cref{eq:ml_head}:
\begin{align}
    \ell_{\rm BCE}(Z^{B}\!\!,Y^{B}\!\!,\mathcal{S}^{B}_{c}) =  - \tfrac{1}{|B|} & \sum\limits_{\mathclap{\substack{i \in B \\ j \in [L]}}} \mathbbm{1}[j=y_{i}] \log(\sigma(z_{ij})) \raisetag{31pt}\\
    - \tfrac{\alpha}{(K-1)|B|}                                                  & \sum\limits_{\mathclap{\substack{i \in B \\ j \in [L]}}} \mathbbm{1}[j{\in}\mathcal{S}_{c}(x_{i})] \log(\sigma(z_{ij}))   \nonumber       \\
    - \tfrac{\alpha}{(L-K)|B|}                                                  & \sum\limits_{\mathclap{\substack{i \in B \\ j \in [L]}}} \mathbbm{1}[j{\notin}\mathcal{S}(x_{i})] \log(1{-}\sigma(z_{ij}))  \nonumber \enspace,
\end{align}
The model is then trained jointly by minimizing the sum of the two losses:
\begin{align}
    \ell_{\rm AVG\text{-}K}(Z'^{B}\!\!, Z^{B}\!\!, Y^{B}) = \ell_{\rm CE}(Z'^{B}\!\!, Y^{B}) + \ell_{ \rm BCE}(Z^{B}\!\!, Y^{B}\!\!, \mathcal{S}^{B}_{c})  \raisetag{-5pt}
\end{align}

At test time, the SCCP head is no longer needed, so we simply use the predictions of the ML head for prediction.
\subsection{Hyperparameters}
\label{subsec:hyperparameters}
Our method depends heavily on the batch size since the candidate classes are selected from the whole batch.
If the batch size is one, then for all $i \in \mathcal{N}_{train}$ $|\mathcal{S}(x_{i})| = K$ with $|\mathcal{S}_{c}(x_{i})| = K -1$ and the method is not able to capture the variability of ambiguity.
As soon as $|B| \geq 2$, the set sizes can vary within the batch, allowing to account for the difference in ambiguity between different images of the batch.
In our experiments, we found that classical values of $|B|$ work well.
The hyperparameter $\alpha$ should be tuned on a validation.
We found that $[0.1, 10.0]$ is a good default search range.
We include experiments in the supplementary material to study the influence of $|B|$ and $\alpha$ on average-$K$ accuracy.
\subsection{Discussion}
\label{subsec:discussion}
Our method has the advantage of being both computationally and memory efficient since it only requires the addition of a linear layer.
In particular, it does not require storing a matrix of size $n_{train} \times L$ in memory as in \cite{cole2021}, which is prohibitive for large datasets.
Besides, instead of a constraint on the average value of the predictions \cite{cole2021}, which can be satisfied by the model in various ways, we dynamically infer the instances target sets with the first head.
\begin{algorithm}
    \caption{Computation of the threshold $\lambda_{val}$.}
    \label{alg:lmbda_val}
    \SetKwInOut{Input}{Input}
    \SetKwInOut{Output}{Output}
    \DontPrintSemicolon
    \Input{$n_{val} \coloneqq |\mathcal{N}_{val}|$, $X_{val} = \{x_{i}\}_{i \in \mathcal{N}_{val}}$, batch size $|B|$, model $f_{\theta}$, $K$}

    \Output{$\lambda_{val}$}
    \BlankLine
    Split $X_{val}$ into $\lceil \frac{|B|}{n_{val}} \rceil$ batches $X_{1}$, \dots, $X_{\lceil \frac{|B|}{n_{val}} \rceil}$

    $f_{1}$ $\gets$ $f_{\theta}(X_{1})$ \tcp*[r]{get batch logits}

    $P$ $\gets$ $\varsigma(f_{1})$ \tcp*[r]{apply softmax}

    \For{$i=2$ \KwTo $\lceil \frac{|B|}{n_{val}} \rceil$}{

        $f_{i}$ $\gets$ $f_{\theta}(X_{i})$ \tcp*[r]{get batch logits}

        $p_{i}$ $\gets$ $\varsigma(f_{i})$ \tcp*[r]{apply softmax}

        $P$ $\gets$ $\mathrm{CONCAT}(P, p_{i})$ \tcp*[r]{row axis concat.}
    }
    \tcc{At this point $P$ is a $n_{val} \times L$ matrix}

    $P$ $\gets$ $\mathrm{FLATTEN}(P)$ \tcp*[r]{turn $P$ into a vector}

    $P$ $\gets$ $\mathrm{SORT}(P)$ \tcp*[r]{sort in decreasing order}

    $\lambda_{val}$ $\gets$ $\frac{1}{2}(P_{[K n_{val}]} + P_{[K n_{val} + 1}])$

    \Return{$\lambda_{val}$}
\end{algorithm}
\begin{table*}[t]
    \begin{center}
        \begin{tabular}{|c|c|c|c|c|c|}
            \hline
            $\ell_{\rm CE}$ & $\ell_{\rm AVG\text{-}K}$ & $\ell_{ROLE}$  & $\ell_{AN}$    & $\ell_{EPR}$ & $\ell_{\rm TOP\text{-}K}$ \\
            \hline\hline
            $96.83\pm0.16$  & $\mathbf{97.35}\pm0.06$   & $96.12\pm0.11$ & $96.71\pm0.02$ & $95.88\pm0.05$                  & $96.32\pm0.05$            \\
            \hline
        \end{tabular}
    \end{center}
    \caption{CIFAR-100 test average-5 accuracy, (DenseNet 40-40)}
    \label{tab:cifar100}
\end{table*}
\section{Experiments}
\label{sec:experiments}
\subsection{Metrics}
\label{subsec:metrics}
We compare $\ell_{\rm AVG\text{-}K}$ with $\ell_{\rm CE}$, $\ell_{\rm AN}$, $\ell_{\rm EPR}$ and $\ell_{\rm ROLE}$ on two datasets with different degrees of ambiguity.
We also include the balanced top-$K$ loss $\ell_{\rm TOP\text{-}K}$ from \cite{topk_icml2022}.
For all datasets, we follow the same workflow: we train a neural network on the dataset with the different methods we compare.
Early stopping is performed on best validation average-$K$ accuracy, computed as follows:
\begin{align}
    \textit{val avg-$K$ accuracy} = \frac{1}{|\mathcal{N}_{val}|} \sum\limits_{i \in \mathcal{N}_{val}} \mathbbm{1}[\varsigma_{y_{i}}(z_{i}) \geq \lambda_{val}] \enspace, \raisetag{6pt}
\end{align}
where the computation of $\lambda_{val}$ is described in \Cref{alg:lmbda_val}.
We then report average-$K$ accuracies on the test set, using the threshold $\lambda_{val}$:
\begin{align}
    \textit{test avg-$K$ accuracy} = \frac{1}{|\mathcal{N}_{test}|} \sum\limits_{i \in \mathcal{N}_{test}} \mathbbm{1}[\varsigma_{y_{i}}(z_{i}) \geq \lambda_{val}] \enspace. \raisetag{6pt}
\end{align}
The hyperparameters specific to all tested methods are tuned on the validation set using grid search, and the best model is then evaluated on the test set.
The results are the average of several runs with different seeds and reported with 95\% confidence intervals.
\subsection{CIFAR100}
\label{subsec:cifar_100}
\textbf{Training}: We first experiment our method on CIFAR-100 \cite{cifar100}, a dataset with $L=100$ classes.
We split the original training set (\num{50000} images)
into a balanced training set of \num{45000} images and a balanced validation set of \num{5000} images on which all hyperparameters are tuned.
We train a DenseNet40-40 \cite{densenet} for 300 epochs with SGD and a Nesterov momentum of 0.9, following \cite{topk_icml2022, berrada}.
The batch size is set to 64 and the weight decay to 0.0001.
The learning rate is initially set to 0.1 and divided by ten at epoch 150 and 225.
\\

\textbf{Results}: We report test average-$5$ accuracy in \Cref{tab:cifar100}.
CIFAR-100 is composed of 20 superclasses each containing 5 classes, \eg the superclass ``aquatic mammals" groups ``beaver", ``dolphin", ``otter", ``seal", and ``whale".
Therefore, most of the ambiguity resides within each superclass, and we are able to achieve high average-5 accuracies ($\sim 96{\text-}97 \%$, cf. \Cref{tab:cifar100}.)
This relatively low ambiguity explains the good performances of $\ell_{\rm CE}$ and $\ell_{AN}$.
We find that $\ell_{EPR}$ and $\ell_{ROLE}$ lag behind, while $\ell_{\rm AVG\text{-}K}$ based on the proposal of candidate sets benefits from a performance gain over all the other methods.

\subsection{Pl@ntNet-300K}
\label{subsec:plantnet_300k}
\begin{table*}
    \begin{center}
        \begin{adjustbox}{width=0.99\textwidth}
            \begin{normalsize}
                \begin{tabular}{|c||c|c|c|c|}
                    \hline
                    $K$                       & 2                                                       & 3                                                                        & 5                                                                        & 10                                                                       \\
                    \hline
                    $\ell_{\rm CE}$           & $89.63 \pm 0.08$  (\textbf{27.08}/\textbf{65.83}/85.97) & $92.64 \pm 0.17$ (38.44/75.55/90.50)                                     & $95.11 \pm 0.18$ (35.39/83.65/94.59)                                     & $97.11 \pm 0.09$ (46.58/91.71/97.30)                                     \\
                    $\ell_{\rm TOP\text{-}K}$ & $\textbf{90.48} \pm 0.05$ (24.98/63.08/\textbf{87.21})  & $93.60 \pm 0.09$ (38.46/73.67/91.85)                                     & $95.75 \pm 0.07$  (49.90/81.39/95.00)                                    & $97.26 \pm 0.03$ (56.49/88.78/97.26)                                     \\
                    $\ell_{\rm AVG\text{-}K}$ & $90.34 \pm 0.06$          (23.77/61.81/86.74)           & $\textbf{93.81} \pm 0.10$ (\textbf{40.39}/\textbf{76.50}/\textbf{92.17}) & $\textbf{96.42} \pm 0.09$ (\textbf{55.83}/\textbf{88.19}/\textbf{95.90}) & $\textbf{98.23} \pm 0.03$ (\textbf{75.75}/\textbf{94.14}/\textbf{98.08}) \\
                    $\ell_{AN}$               & $85.41 \pm 0.15$          (2.05/30.26/76.64)            & $90.15 \pm 0.17$ (7.46/47.20/86.04)                                      & $93.88 \pm 0.12$ (20.31/67.55/92.66)                                     & $96.86 \pm 0.06$ (42.61/85.49/96.87)                                     \\
                    $\ell_{\rm EPR}$          & $86.30 \pm 0.17$          (9.01/37.27/77.72)            & $90.49 \pm 0.14$ (18.49/51.58/85.19)                                     & $93.63 \pm 0.04$ (31.02/65.85/90.79)                                     & $95.99 \pm 0.04$ (41.44/78.61/95.09)                                     \\

                    \hline
                \end{tabular}
            \end{normalsize}
        \end{adjustbox}
    \end{center}
    \caption{Pl@ntNet-300K test average-$K$ accuracy (ResNet-18). The three numbers in parentheses represent respectively the mean average-$K$ accuracies of 1) few shot classes ($<20$ training images) 2) medium shot classes ($20 \leq \cdot \leq 100$ training images) 3) many shot classes ($> 100$ training images).}
    \label{tab:plantnet300k}
\end{table*}
\textbf{Description}: Pl@ntNet-300K \cite{plantnet-300k} is a plant image dataset of \num{306146} images and \num{1081} classes (species) that contains a lot of class ambiguity.
It is composed of 303 \textit{genera} (superclasses) each comprising one or more species.
Ambiguity is particularly important within a \textit{genus}: for instance, two orchid species may be very similar.
In Pl@ntNet-300K, a \textit{genus} can include from one to several dozen species.
This makes the ambiguity in this dataset very variable (see \Cref{fig:plant_pictures}).
Moreover, the labels are crowdsourced so Pl@ntNet-300K is particularly prone to label noise.
These reasons make it a perfect candidate for average-$K$ classification. \\

\textbf{Training}: We finetune a ResNet-18 \cite{resnet} pre-trained on ImageNet \cite{imagenet} for 30 epochs.
We use SGD and a Nesterov momentum of 0.9, with a batch size of 32 and an initial learning rate of 0.002, divided by ten at epoch 25.
The weight decay is fixed at 0.0001. All methods are trained for $K \in \{2, 3, 5, 10\}$ and the average-$K$ accuracy on the test set is then reported for each value of $K$.\\
\begin{figure*}[t]
    \centering
    \begin{subfigure}[b]{0.48\textwidth}
        \centering
        \includegraphics[width=\textwidth]{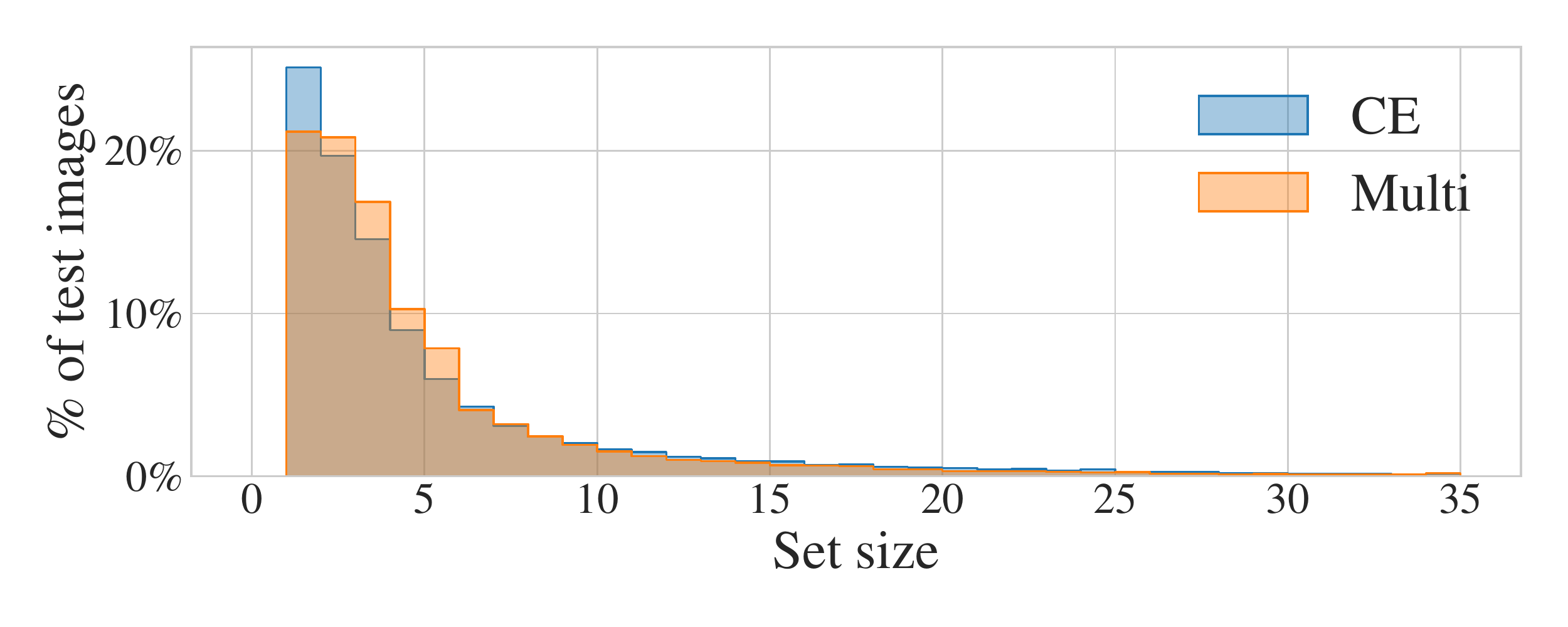}
        \caption{Histogram of set sizes for Pl@ntNet-300K test set images for a model trained with $\ell_{\rm CE}$ or $\ell_{\rm AVG\text{-}5}$ to return sets of size 5 on average (ResNet-18)}
        \label{subfig:set_size}
    \end{subfigure}
    \hfill
    \begin{subfigure}[b]{0.48\textwidth}
        \centering
        \includegraphics[width=\textwidth]{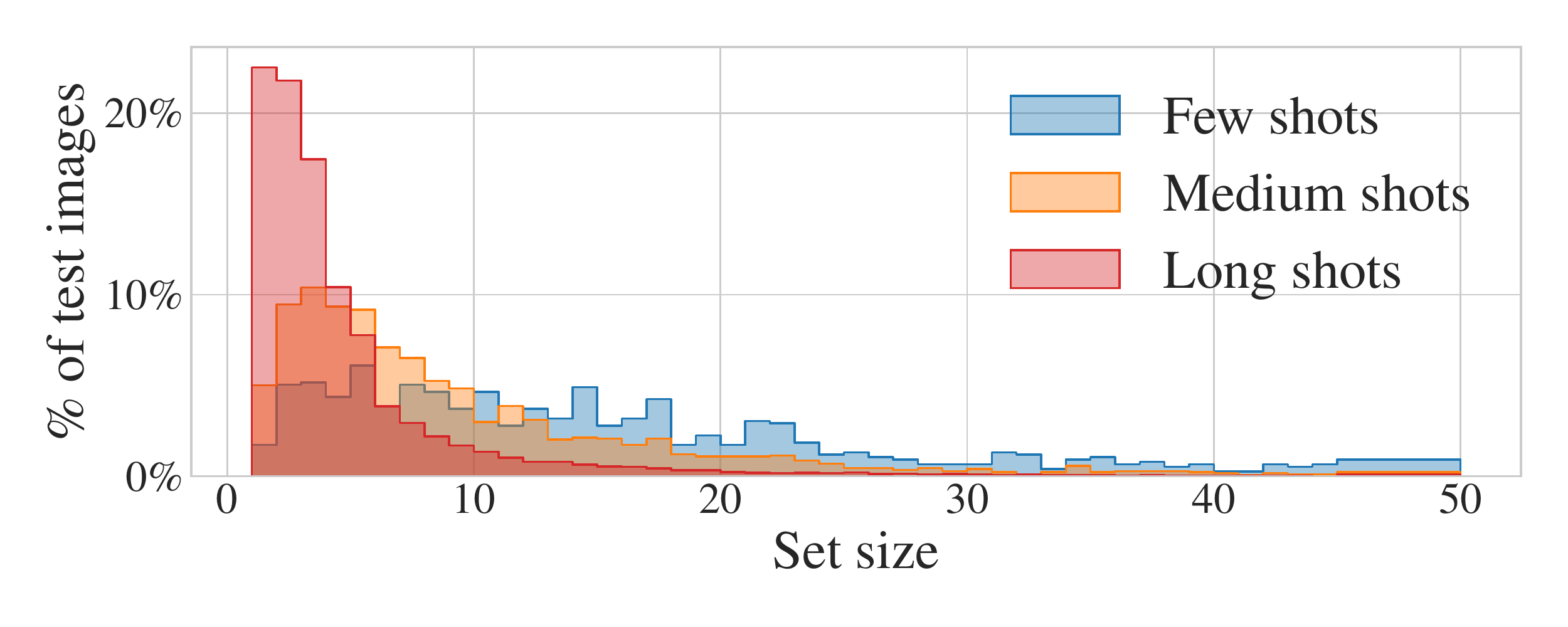}
        \caption{Histogram of set sizes for Pl@ntNet-300K test set images by category (few-shot, medium shot, long-shot classes) for a model trained with $\ell_{\rm AVG\text{-}5}$ to return sets of size 5 on average (ResNet-18)}
        \label{subfig:three_categories_multi}
    \end{subfigure}
    \caption{Histograms of set sizes.}
    \label{fig:toname}
\end{figure*}
\textbf{Results and interpretation}: The results can be found in \Cref{tab:plantnet300k}. It shows that our method is more effective for all $K$ except when $K = 2$ where it gives results similar to the top-$K$ loss.
For high values of $K$, the gain is particularly important for few-shot and medium-shot classes, \ie classes with few examples (for a precise definition of few/medium/long shot classes, see the caption of \Cref{tab:plantnet300k}).
For instance, for $K =10$, the average-10 accuracy of few-shot classes is $75.75$ for our method compared to $56.49$ for the top-$K$ loss and $46.58$ for the cross-entropy.
These results are interesting because of Pl@ntNet-300K's long-tail: few-shot and medium-shot classes account for 48\% and 25\% of the total number of classes, respectively.
This means that the model recognition capabilities are significantly higher for a vast majority of the classes.
It should be noted that high values of $K$ can arise in certain applications, for instance for the diagnosis of rare diseases \cite{faviez2020diagnosis} or for the automatic prediction of likely species at a particular location \cite{lorieul2022overview}.

From \Cref{tab:plantnet300k} we see that the most naive methods $\ell_{AN}$ and $\ell_{\rm EPR}$ perform poorly compared to the other losses.
In particular, $\ell_{AN}$ which gave decent performances on CIFAR-100 struggles on Pl@ntNet-300K.
One possible reason is that $\ell_{AN}$ assumes no class ambiguity while Pl@ntNet-300K has many.

It is worth noting that $\ell_{\rm TOP\text{-}K}$ gives interesting results and even outperforms cross-entropy.
This is not surprising since the top-$K$ loss optimizes top-$K$ accuracy, and average-$K$ and top-$K$ classifiers are somewhat related (top-$K$ classifiers are particular average-$K$ classifiers that return $K$ classes for each example).
For a thorough comparison of top-$K$ and average-$K$ classification, we refer the reader to \cite{lorieulthese}. \\

\textbf{Distribution of set sizes}: \Cref{subfig:set_size} shows the repartition of set sizes for Pl@ntNet-300K test images, for a model trained with $\ell_{\rm AVG\text{-}5}$ or $\ell_{\rm CE}$ to return sets of average size 5.
The cross-entropy is over-confident for many images, which results in an important number of sets of size one, whereas our method is more conservative and tends to return fewer sets of size one but more sets of size two, three and four.

\Cref{subfig:three_categories_multi} shows the distribution of set size for few-shot, medium-shot and long-shot classes for a model trained with $\ell_{\rm AVG\text{-}5}$ to return sets of average size 5.
It appears clearly that images belonging to long-shot classes are associated with small set sizes.
This is because the model saw enough training images belonging to these classes to make confident predictions.
For medium-shot classes, the mode of the distribution is higher and the tail is heavier.
For the most challenging images that belong to few-shot classes, the uncertainty results in larger set sizes, going up to $\sim 50$ classes.
\section{Limitations and future work}
\label{sec:limitations}
While $\ell_{\rm AVG\text{-}K}$ allows practical gains in average-$K$ accuracy,
its particular structure, based on a two-headed deep neural network makes its theoretical analysis difficult.
In particular, it can not be shown easily that $\ell_{\rm AVG\text{-}K}$ is average-$K$ calibrated \cite{lorieulthese} like the cross-entropy.

We have proposed a loss function for average-$K$ classification which is a particular instance of set-valued classification.
Other settings exist \cite{chzhen2021set} (point-wise error control, average coverage control) and could be the object of ad-hoc optimization methods in future work.


\section{Conclusion}
\label{sec:conclusion}
We propose a loss function to optimize average-$K$ accuracy, a setting in which sets of variable size are returned by the classifier to reduce the risk.
Our method is based on the addition of an auxiliary head in the deep neural network trained the cross-entropy whose goal is to propose candidate class sets for the current batch of images.
The candidate classes identified by the auxiliary head are then treated as pseudo-positives by a multi-label head optimized with the binary cross entropy.
We show that our method compares favorably to the cross-entropy loss and other binary methods as well as to a top-$K$ loss.
We further show that the gain in average-$K$ accuracy increases with $K$ and is substantial for classes in the tail of a heavily imbalanced dataset.
Our method has the advantage to be both memory and computationally efficient since it estimates the candidate classes on the fly with a single linear layer.

\section*{Acknowledgements}
This work was funded by the French National Research Agency (ANR) through the grant CaMeLOt ANR-20-CHIA-0001-01 and the grant Pl@ntAgroEco 22-PEAE-0009. The authors are grateful to the OPAL infrastructure from Université Côte d'Azur for providing resources and support.



{\small
    \bibliographystyle{ieee_fullname}
    \bibliography{../../sample}
}

\newpage
\appendix
\onecolumn

\section{Hyperparameter sensitivity}
\label{sec:hyper_sensitivity}
\vspace{1cm}
We study the impact of the hyperparameters $\alpha$ and $|B|$ on average-$K$ accuracy by conducting further experiments on CIFAR-100 (\Cref{subsec:cifar_100}).

\vspace{1cm}

\begin{figure}[h]
    \centering
    \includegraphics[width=0.65\linewidth]{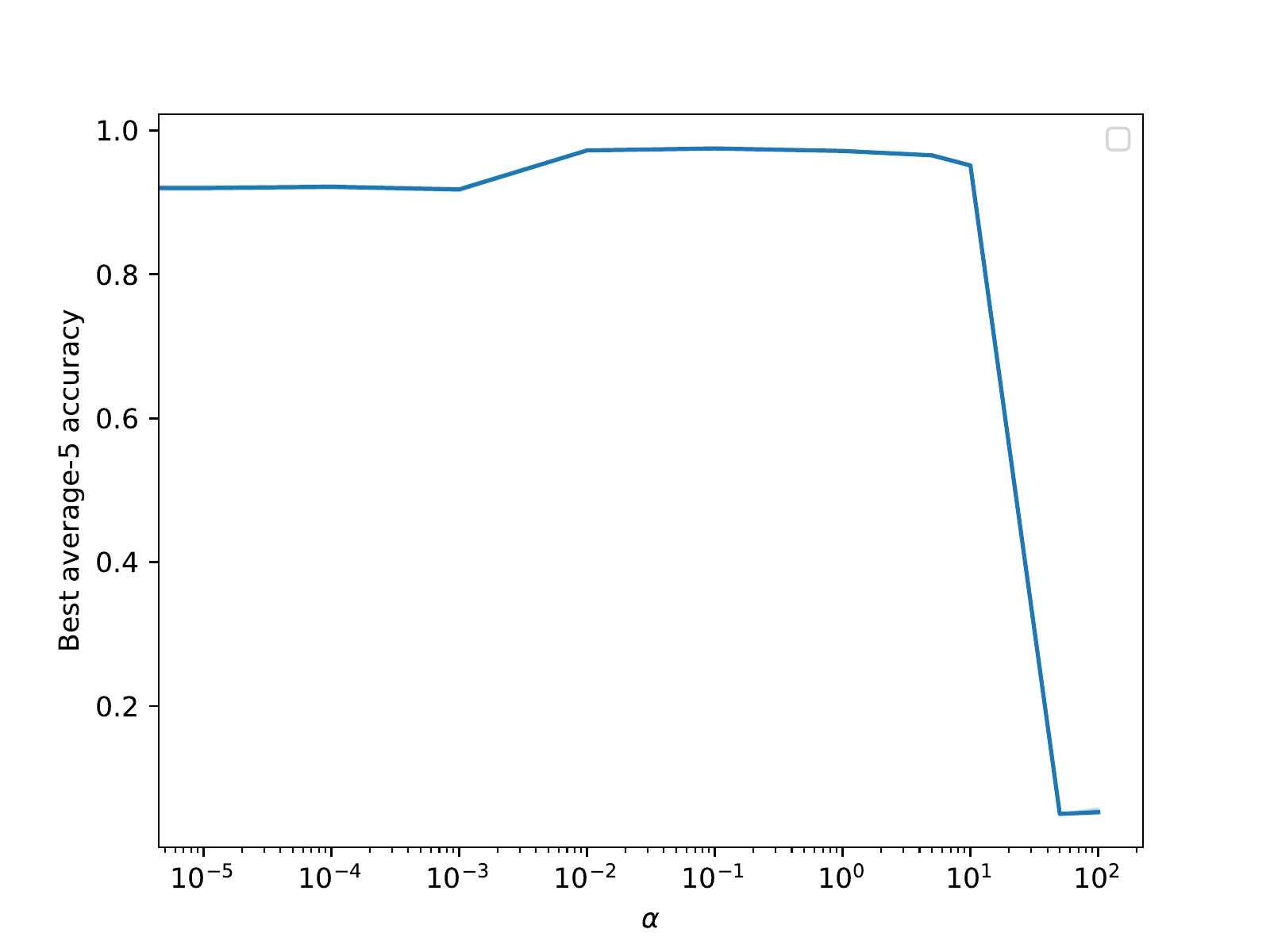}
    \caption{CIFAR-100 best validation average-5 accuracy for a DenseNet 40-40 trained with $\ell_{\rm AVG\text{-}5}$ for different values of $\alpha$.}
    \label{fig:plot_alpha}
\end{figure}

\Cref{fig:plot_alpha} shows how CIFAR-100 average-5 accuracy varies as a function of the hyperparameter $\alpha$.
Average-5 accuracy is stable over a wide range of $\alpha$ values (roughly $10^{-2}$ to $10^{1}$), which means that $\alpha$ does not require a precise tuning to obtain good results.
It drops sharply for high $\alpha$ values, \ie when the candidate classes have much more weight than the annotated labels of the training set in the objective function.

\vspace{1cm}

\begin{figure}[t]
    \centering
    \includegraphics[width=0.6\linewidth]{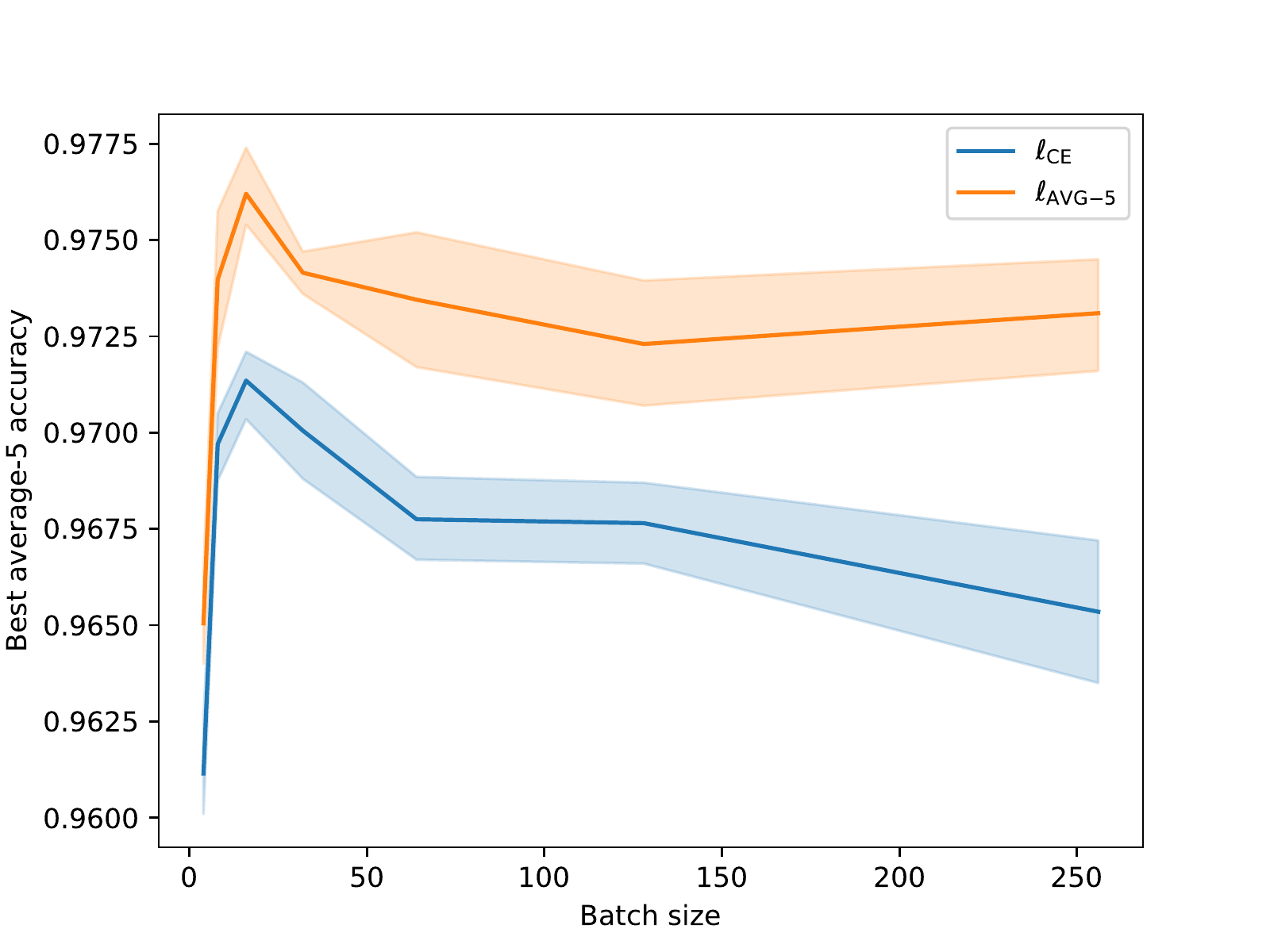}
    \caption{CIFAR-100 best validation average-5 accuracy as a function of batch size for a DenseNet 40-40 trained with $\ell_{\rm AVG\text{-}5}$ or $\ell_{\rm CE}$.
    For a fair comparison we maintain the ratio of learning rate to batch size constant.
    The 95\% confidence interval is represented.}
    \label{fig:bs}
\end{figure}

\Cref{fig:bs} shows how average-$K$ accuracy varies with the batch size for a model trained with $\ell_{\rm CE}$ or $\ell_{\rm AVG\text{-}K}$.
For a fair comparison we maintain the ratio of learning rate to batch size constant.
As expected, average-$K$ accuracy decreases for both methods when the batch size becomes too small.
However, we find that $\ell_{\rm AVG\text{-}K}$ is more robust than $\ell_{\rm CE}$ to large batch size values.
This can be explained by the choice of more relevant candidate classes when the batch size becomes large.
This is counterbalanced by the empricical fact that SGD tends not to work well with very large batch sizes in deep learning.

\section{Experiments details}
\label{sec:exp_details}
\vspace{1cm}
We report in \Cref{tab:hyperparameters_cifar,tab:hyperparameters_plantnet} the hyperparameters selected after grid search for all losses,
for CIFAR-100 and Pl@ntNet-300K datasets respectively.

\begin{table}[h]
    \begin{center}
        \begin{tabular}{|c|c|}
            \hline
            loss                          & hyperparameters                  \\
            \hline\hline
            $\ell_{\rm CE}$               &         -                        \\
            $\ell_{\rm AVG\text{-}K}$     &        $\alpha =0.3$             \\
            $\ell_{AN}$                   &          -                        \\
            $\ell_{EPR}$                  &          $\beta = 0.01$                  \\
            $\ell_{\rm TOP\text{-}K}$     &          $\epsilon = 0.2$                        \\
            $\ell_{ROLE}$                 &          $\lambda = 0.0$, learning rate $\Theta$: $\times 1.0$                 \\
            \hline
        \end{tabular}
    \end{center}
    \caption{Hyperparameters selected after grid search for the CIFAR-100 experiments.}
    \label{tab:hyperparameters_cifar}
\end{table}

\begin{table}[t]
    \begin{center}
        \begin{tabular}{|c|c|}
            \hline
            loss                          & hyperparameters                  \\
            \hline\hline
            $\ell_{\rm CE}$               &         -                        \\
            $\ell_{\rm AVG\text{-}K}$     &        $\alpha =5.0$             \\
            $\ell_{AN}$                   &          -                        \\
            $\ell_{EPR}$                  &          $\beta = 0.001$                  \\
            $\ell_{\rm TOP\text{-}K}$     &          $\epsilon = 1.0$                        \\
            \hline
        \end{tabular}
    \end{center}
    \caption{Hyperparameters selected after grid search for the Pl@ntNet-300K experiments.}
    \label{tab:hyperparameters_plantnet}
\end{table}

\end{document}